\documentclass[letterpaper]{article}
\usepackage{spconf,amsmath,graphicx, amsfonts}
\usepackage{epstopdf}
\usepackage[lined,boxed]{algorithm2e}
 \pdfoutput=1
\usepackage[english]{babel}
\usepackage{fancyhdr}
\usepackage{threeparttable}
\usepackage{parskip}
\usepackage{cite}
 \usepackage{hyperref}
\usepackage{caption}
\usepackage{subfig}
\usepackage{float}
\usepackage{graphicx}
\usepackage{tablefootnote}
\usepackage{booktabs}
\usepackage{multirow}
\usepackage[T1]{fontenc}
\usepackage{pgf}
\usepackage[normalem]{ulem}
\usepackage[inline]{enumitem}
\usepackage[autoplay]{animate}

\newcommand{\be}{\begin{equatio}}

\newcommand{\norm}[1]{\left\|#1\right\|}

\def\bmt{\left(\begin{matrix}}

\def\emt{\end{matrix}\right)}

\def\by{\mathbf{y}}

\def\bs{\mathbf{s}}
\def\bX{\mathbf{Y}}
\def\bY{\mathbf{Y}}
\def\bD{\mathbf{D}}

\def\bM{\mathbf{M}}
\def\bS{\mathbf{S}}
\def\be{\mathbf{e}}
\def\bE{\mathbf{E}}

\def\Fb{\mathbf{f}}
\def\Fb{\mathbf{F}}

\def\barX{\bar{\mathbf{Y}}}

\def\barS{\bar{\mathbf{S}}}

\def\barN{\bar{N}}

\def\trace{\textrm{trace}}
\def\etal{\textit{et al.}}

\def\R{\mathbb{R}}

\title{DFDL: DISCRIMINATIVE FEATURE-ORIENTED DICTIONARY LEARNING \\FOR HISTOPATHOLOGICAL IMAGE CLASSIFICATION}
\twoauthors
  {Tiep H. Vu, Hojjat S. Mousavi, Vishal Monga\vspace{-.2in} \thanks{\hspace{-1.7em}Research was supported by an Army Research Office Grant Number \textbf{W911NF-14-1-0421} }}
	{\small Pennsylvania State University, University Park, PA}
  {UK Arvind Rao, Ganesh Rao\vspace{-.2in}}
	{\small UT MD Anderson Cancer Center, Houston, TX}

\begin{document}
\maketitle
\thispagestyle{fancy}
\pagestyle{empty}
\begin{abstract}
\vspace{0.1in}
\label{abstract}
In histopathological image analysis, feature extraction for classification is a challenging task due to the diversity of histology features suitable for each problem as well as presence of rich geometrical structure. In this paper, we propose an automatic feature discovery framework for extracting discriminative class-specific features and present a low-complexity method for classification and disease grading in histopathology. Essentially, our Discriminative Feature-oriented Dictionary Learning (DFDL) method learns class-specific features which are suitable for  representing samples from the same class while are poorly capable of representing samples from other classes. Experiments on three challenging real-world image databases: 1) histopathological images of intraductal breast lesions, 2) mammalian lung images provided by the Animal Diagnostics Lab (ADL) at Pennsylvania State University, and 3) brain tumor images from The Cancer Genome Atlas (TCGA) database, show the significance of DFDL model in a variety problems over state-of-the-art methods.
\end{abstract}
\begin{keywords}
Histopathological image classification, Sparse coding, Dictionary learning, Feature extraction
\end{keywords}

\section{Introduction}
\label{sec:intro}
 Automated histopathological image analysis has recently become a significant research problem in medical imaging and there is an increasing need for developing quantitative image analysis methods as a complement to the effort of pathologists in diagnosis process.
Consequently, an emerging class of problems in medical imaging focuses on the the development of computerized frameworks to classify histopathological images \cite{Gurcan2009,Srinivas2013,Srinivas2014SHIRC,Nandita2013,Mousavi2015JPI}. These advanced image analysis methods have been developed with purpose of relieving the workload on pathologists by sieving out obviously diseased and also healthy cases, which allows specialists to spend more time on more sophisticated cases.
\par
In the diagnosis process, pathologists often look for problem-specific visual cues in histopathological images in order to categorize a tissue image as one of the possible categories. Consequently, different customized feature extraction techniques for a variety of problems have been developed based on these visual cues \cite{Orlov2008,Shamir2008,Gultekin2014,Shi2014,Minaee2013prediction}. However, a challenging question in medical image analysis is how to extract these features. The challenge inherits from the richness of geometric structures in tissue imagery and the meaningful pathological information at diverse scales. Although several methods have been proposed for this crucial task, they are mostly exclusively designed for particular data sets and are highly dependent on preprocessing steps (e.g., color normalization and nuclear segmentation), limiting their performance on general histopathology problems. In order to mitigate the workload in preprocessing step and to develop a more general solution, we propose a dictionary learning method relying on a sparse representation-based framework that can automatically discover relevant features from raw medical images and can be applied to several histopathological data sets.
\par
{Sparse representation-based methods are powerful tools for image classification \cite{Wright2009SRC,Bahrampour2014,Mousavi2014ICIP}. The underlying idea is that given a class of images and sufficient collection of bases, a test image can be expressed approximately as a sparse linear combination of bases. Representing signals using a set of learned bases instead of predefined bases, e.g. DCT  and wavelet bases, has led to state-of-the-art results in various applications such as denoising, inpainting and classification \cite{elad2006image,Aharon2006KSVD,mairal2009online}. To achieve a comprehensive set of bases, sparsity and task-driven constraints are combined together in several ways into optimization problems, which are called Dictionary Learning methods.} For classification problems, the class-specific design of such dictionaries enables class assignment via a simple reconstruction error-based metric \cite{Anaraki2013,Sadeghi2014learning}. In particular, GDDL \cite{Suo2014} and LC-KSVD \cite{Zhuolin2013LCKSVD}  enforced the label consistency needed between dictionary bases and training data for classification. Meanwhile, FDDL \cite{Meng2011FDDL} encouraged coding coefficients to have small intra-class scatter but big inter-class scatter.

\par
{Sparsity-based} classification schemes have also been proposed for medical applications, recently \cite{Liu2011,Zhang2010}. Specifically, Srinivas \etal \cite{Srinivas2013,Srinivas2014SHIRC} presented a multi-channel histopathological image as a sparse linear combination of training examples under channel-wise constraints and proposed a residual-based classification technique. In addition, Parvin \etal \cite{Nandita2013} combined a dictionary learning framework with a Restricted Boltzmann Machine to learn sparse features for classification.
\par
\begin{figure}[t]
\centering
  \includegraphics[width=0.45\textwidth]{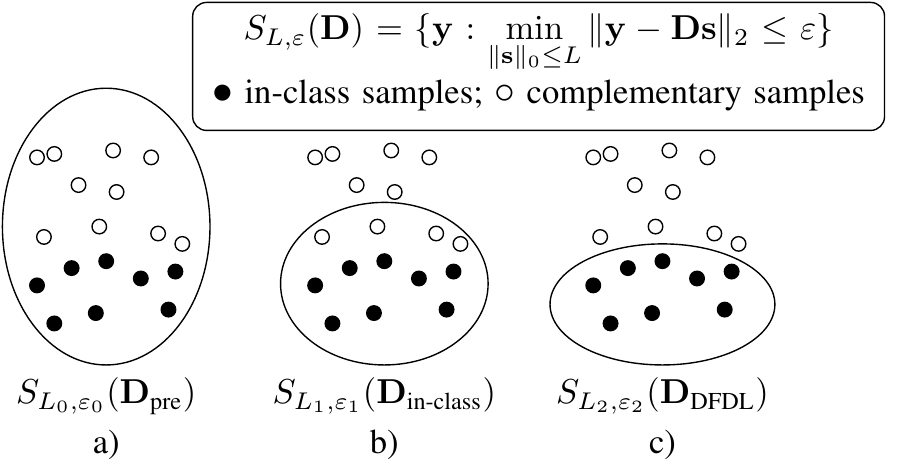}
  \vspace{-0.1in}
 \caption{{\small Sparse representation space of a) predefined dictionary, e.g. DCT or Wavelet ($S_{L_0,\varepsilon_0}(\bD_{\text{pre}})$ may cover all data), b) learned dictionary using in-class samples only, e.g. KSVD\cite{Aharon2006KSVD} or ODL\cite{mairal2009online} ($S_{L_1,\varepsilon_1}(\bD_{\text{in-class}})$ may also cover some complementary samples) and c) desired DFDL ($S_{L_2,\varepsilon_2}(\bD_{\text{DFDL}})$ cover in-class samples only).}}
  \label{fig: idea}
\end{figure}

Being mindful of the challenges of feature extraction of histopathological images, we aim to build discriminative bases for each class by imposing sparsity constraints on minimizing intra-class differences, {while simultaneously} emphasizing inter-class differences. Small intra-class differences encourage the comprehensibility of the set of learned bases, which has the ability of representing in-class samples with only few bases (intra class sparsity). Simultaneously, large inter-class differences prevent bases of a class from sparsely representing samples from other classes (complementary samples).  This crucial property of learned bases would promote the discrimination {ability of the sparse code (coefficient vector) for classification}. Concretely, given a dictionary from a particular class $\bD$ containing $k$ bases and a certain number $L\ll k$, we define an \emph{$L$-subspace} of $\bD$ as a span of a subset of $L$ bases from $\bD$. Our proposed Discriminative Feature-oriented Dictionary Learning (DFDL) aims to build dictionaries with this key property: any sample from a class can be reasonably close to an $L$-subspace of the associated dictionary while a complementary sample is far from \emph{any} $L$-subspace of that dictionary. Illustration of the proposed idea is shown in Fig. \ref{fig: idea}.

\par
\textbf{Contributions:} The main contributions of this paper are as {follows: (1) A} dictionary learning method for automatic feature discovery in histopathological images to mitigate the generally difficult problem of feature extraction in medical images. (2) Our framework is a \emph{discriminative} dictionary learning method that emphasizes inter-class differences while keeping intra-class differences small, resulting in {enhanced} classification performance. (3) The proposed method is applied on three diverse histopathological data sets to show the capability of our method in handling {a variety of diagnosis and grading problems}. Extensive experimental results show that our method provides outstanding performances even with a small number of training images.
\vspace{-.15in}
\section{Discriminative dictionary learning}
\vspace{-0.05in}
\subsection{Notation} 
\vspace{-0.05in}
\label{sub:notaions}
Suppose that we have $c$ classes. The vectorization of a small block (or patch) of an image\footnote{ {Currently, we are working on the luminance component only, extension of DFDL to multi-channel case will be considered in future.}}, which will be referred as a sample, is denoted as a column vector $\by \in \R^d$. For $i = 1, 2, \dots, c$, let $\bX_i \in \R^{d\times N_i}$ and $\bar{\bX}_i \in \R^{d \times \bar{N}_i}$ be matrices containing all data samples from class $i$ and its complementary samples, respectively. We denote by $\bD_i \in \R^{d \times k_i}$ the dictionary of class $i$. 
\par
For a code $\bs \in \R^k$, we denote by $\|\bs\|_0$ the number of its non-zeros. The sparsity constraint of $\bs$ can be formulated as $\|\bs\|_0 \le L$ with $L \ll k$. For a matrix $\bS$ , $\|\bS\|_0 \le L$ means that \emph{each} column of $\bS$ has no more than $L$ non-zeros.

\subsection{Problem Formulation} 
\label{sub:problem_formulation}
We aim to build {class-specific} dictionaries $\bD_i$ such that each $\bD_i$ \emph{reasonably} represents samples from class $i$ but is \emph{poorly} capable of representing its complementary samples. Concretely, for the learned dictionaries we need:
 \begin{eqnarray*}
 & \displaystyle \frac{ 1  } { N_i } \min_{   \|\bS_i\|_0 \leq L_i }\|\bX_i - \bD_i \bS_i\|_F^2 & \text{to be small}\\
 \text{and } & \displaystyle \frac{ 1  } { \bar{N_i} } \min_{\|\bar{\bS}_i\| \leq L_i}\|\bar{\bX}_i - \bD_i \bar{\bS}_i\|_F^2 & \text{to be large}
 \end{eqnarray*}
 where $L_i$ controls the sparsity level and  $\| \bullet\|_F$ denotes the Frobenius norm. For simplicity, from now on, we consider only one class and drop the class index in each notion, i.e., using $\bX, \bD, \bS, \bar{\bS}, N, \bar{N}, L$ instead of $\bX_i, \bD_i, \bS_i, \bar{\bS}_i, N_i, \bar{N}_i$ and $L_i$.
 Based on the argument above, we formulate the optimization problem for each dictionary:
 \begin{align}
 \label{eqn:findDopt}
   \bD^*= \arg \min _{\bD} \Big(&\frac{ 1  } { N }\min_{\|\bS\|_0 \le L}\|\bX - \bD \bS\|_F^2 \nonumber\\
   &- \frac{ \rho  } { \bar{N} } \min_{\|\bar{\bS}\|_0 \le L}\|\bar{\bX} - \bD \bar{\bS}\|_F^2 \Big)
 \end{align}
 where $\rho$ is a positive regularization parameter. The first term in the above optimization problem minimizes intra-class differences and the second term emphasizes inter-class differences. By solving the above problem, we can find the appropriate dictionaries as we desire.
 \par
In the same manner with SRC \cite{Wright2009SRC}, a new patch $\by$ is classified as follows. Firstly, the sparse codes $\hat{\bs}$ are calculated via $l_1$-norm minimization:
    $$\hat{\bs} = \arg \min_{\bs} \big\{ \|\by - \bD_{total}\bs\|_2^2 + \lambda \|\bs\|_1 \big\}$$
    where $\bD_{total} = [\bD_1, \bD_2, \dots, \bD_c]$ is the collection of all dictionaries and $\lambda$ is a scalar constant.
    Secondly, the identity of $\by$ is determined as: $\displaystyle \arg \min_{1 \le i\le c}\{\delta_i(\by)\} $ where
    $$\delta_i(\by) = \|\by-\bD_i \delta_i(\hat{\bs})\|$$
    and $\delta_i(\hat{\bs})$ is part of $\hat{\bs}$ associated with class $i$.

\subsection{Proposed solution }
\label{subsec: solution}
We use an iterative method to find the optimal solution for problem (\ref{eqn:findDopt}). Specifically, the process is iterative by fixing $\bD$ while optimizing $\bS, \bar{\bS}$ and vice versa. 
\par
At sparse coding step, $\bS^*, \bar{\bS}^*$ can be found by solving:
$$
       \bS^* = \arg \min_{\|\bS\|_0 \le L}\|\bX - \bD\bS\|_F^2;~~
       \bar{\bS}^* = \arg \min_{\|\bar{\bS}\|_0 \le L} \|\bar{\bX} - \bD\bar{\bS}\|_F^2
       $$
\par  With the same dictionary, these two sparse coding problems can be combined into the following one:
  \begin{equation}\label{eqn:findS}
    \hat{\bS}^* = \arg\min_{\|\hat{\bS}\|_0 \le L}\norm{\hat{\bX} - \bD\hat{\bS}}_F^2
  \end{equation}
  with $\hat{\bX} = [\bX, \bar{\bX}]$ being the matrix of all training samples and $\hat{\bS} = [\bS, \bar{\bS}]$. This sparse coding problem can be solved by OMP method \cite{tropp2007signal} using SPAMS toolbox \cite{SPAMS}.
  \par
For the bases update stage, $\bD^*$ is found by solving: 
\begin{eqnarray}
	\bD^* &= \arg\min_{\bD} \Big\{ \frac{ 1  } { N } \|\bX - \bD\bS\|_F^2 - \frac{ \rho } { \barN } \|\bar{\bX} - \bD\bar{\bS}\|_F^2 \Big\} \label{subeqn:findD1}\\
	&= \arg\min_{\bD} \big\{-2\trace(\bE \bD^T) + \trace(\bD \Fb \bD^T) \label{subeqn:findD2} \big\}
\end{eqnarray}
\noindent
using the method of block coordinate descent with a warm start to update bases one by one \cite{mairal2009online}. We have used the equation $\|\bM\|_F^2 = \trace(\bM\bM^T)$ to derive (\ref{subeqn:findD2}) from (\ref{subeqn:findD1}) and denoted
      $ \displaystyle
        \bE = \frac{ 1 } { N } \bX \bS^T - \frac{ \rho } { \barN } \barX\barS^T; \quad
        \Fb = \frac{ 1  } { N } \bS \bS^T - \frac{ \rho } { \barN } \barS \barS^T.
        $
\par
	\begin{algorithm}[t]
	 \KwData{$\bY, \bar{\bY}$: collection of all in-class samples and complementary samples. \par
	 $k$: number of learned bases.\\ $\rho$: regularization parameter. $L$: sparsity level}
	 \KwResult{$\bD$: dictionary }
	 1. Initializing $\bD$ by randomly picking $k$ columns of $\bY$
	 \While{not converged}{
	 2. Fix $\bD$ and update $\bS, \bar{\bS}$ by solving Problem (\ref{eqn:findS}); \\
	 3. Fix $\bS, \bar{\bS}$, calculate: $ \displaystyle
	         \bE = \frac{ 1 } { N } \bX \bS^T - \frac{ \rho } { \barN } \barX\barS^T; \quad
	         \Fb = \frac{ 1  } { N } \bS \bS^T - \frac{ \rho } { \barN } \barS \barS^T.
	         $
	      \If {$\Fb$ is not PSD}{
	      $\rho \leftarrow 0.9\rho$; \textbf{go to} 3.; }
	 4. Update $\bD$ by solving Problem (\ref{subeqn:findD2});}
	 \caption{DFDL for sparse representation-based classification}
	 \label{alg:11}
	\end{algorithm}
In order to make the optimization problem tractable and solve it efficiently, we need one more requirement for $\rho$ to ensure that $-2\trace(\bE\bD^T) + \trace(\bD^T\bD\Fb)$ is a convex function with respect to $\bD$, or in other words, the symmetric matrix $\Fb$ need to be positive semidefinite (PSD). If we let $\lambda_1(\bM) \le \lambda_2(\bM) \le \dots \le \lambda_{\max}(\bM)$ be (real) eigenvalues of a symmetric matrix $\bM$, the PSD constraint of $\Fb$ is equivalent to the non-negativity constraint of $\lambda_1(\Fb)$. Using Weyl's inequalities, we can get lower bound for $\lambda_1(\Fb)$:$\lambda_0 = \frac{1}{N}\lambda_1(\bS\bS^T) - \frac{\rho}{\bar{N}} \lambda_{\max}(\barS\barS^T) \le \lambda_1(\Fb) $. As a result, if $\rho$ is guaranteed to be small enough such that $\lambda_0 \geq 0$, then $\Fb$ is PSD. In fact, this problem is unstable and difficult to track since $\bS, \bar{\bS}$ depend on $\bD$. We propose a practical solution for this difficulty as follows. First, we start with a small value of $\rho$ and check if $\Fb$ is PSD at each iteration. If so, $\rho$ remains unchanged; otherwise, $\rho$ would be assigned to a smaller value, say, $0.9\rho$. In our experiments, $\rho = 0.001$ gives good results. Our DFDL method is summarized in Algorithm \ref{alg:11}.

\section{Experimental Results} 
\label{sec:experiment_results}
\begin{figure}[t]
\centering
\subfloat[{UDH}]			{\includegraphics[height=0.13\textwidth]{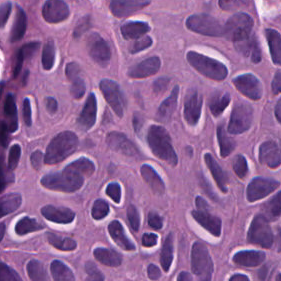}} ~
\subfloat[{Healthy Lung}]	{\includegraphics[height=0.13\textwidth]{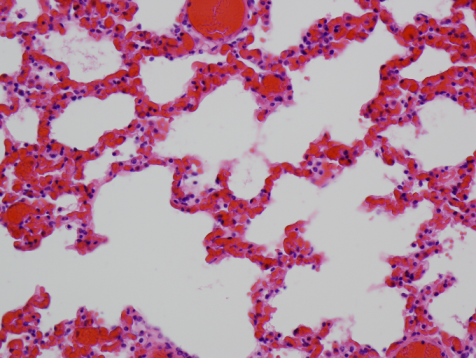}}~
\subfloat[{Not MVP}]		{\includegraphics[height=0.13\textwidth]{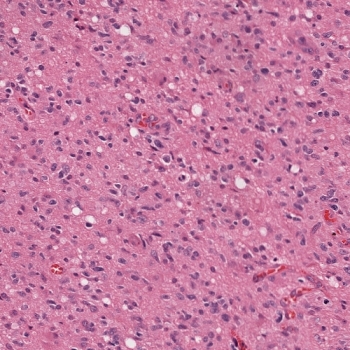}} \\
\subfloat[{ DCIS }  ]		{\includegraphics[height=0.13\textwidth]{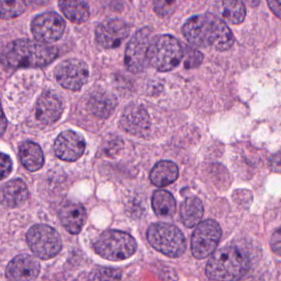}}~
\subfloat[{Inflamed Lung}] 	{\includegraphics[height=0.13\textwidth]{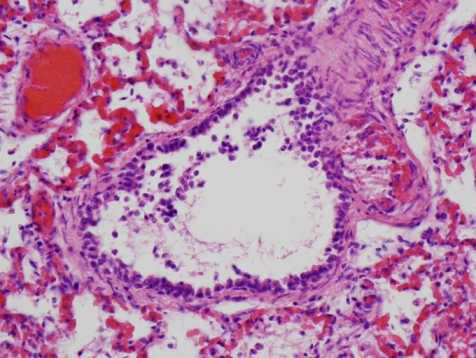}}~
\subfloat[{ MVP}\label{subfig:MVP}]			{\includegraphics[height=0.13\textwidth]{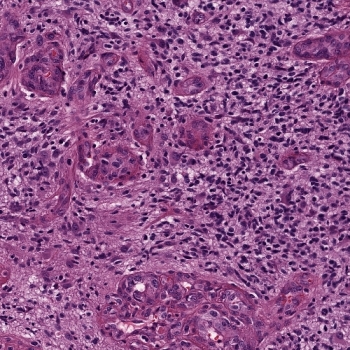}}
\vspace{-0.1in}
 \caption{  Samples from three data sets. Column 1: IBL data set, column 2: ADL data set, column 3: TCGA data set}
 \vspace{-0.05in}

  \label{fig: samples}
\end{figure}

In this section, we present the experimental results of applying DFDL to three diverse histopathological image databases (sample images are shown in Fig. \ref{fig: samples}.) and compare our results with those using {SVM in conjunction with a collection of state-of-the-art histopathology features from WND-CHARM\cite{Shamir2008} (will be referred as WND-CHARM method)}, SRC \cite{Wright2009SRC}, LA-SHIRC \cite{Srinivas2014SHIRC} and other DL methods (LC-KSVD \cite{Zhuolin2013LCKSVD} and FDDL \cite{Meng2011FDDL}). In each experiment,  10000 20-by-20 patches are randomly extracted from training images for each class. Each Dictionary Learning method learns the same number of bases, say 500, per class.

\textbf{IBL data set} contains images which belong to either of two well-defined classes: usual ductal hyperplasia (UDH) and ductal carcinoma in situ (DCIS). Ground truth class labels for the images are assigned manually by the pathologists. A total of 40 patient cases -- 20 well-defined DCIS and 20 UDH -- are identified for experiments in the manner described in \cite{Dundar2011}. Each case contains a number of regions of interest (RoIs), and we have chosen a total of 120 images (RoIs), consisting of a randomly selected set of 20 images for training and the remaining 100 RoIs for test. Images are downsampled for computational reduction purpose such that size of a cell is around 20-by-20 (pixel). 
 In classification step, an image is decomposed into non-overlapping patches and it is classified as Healthy if proportion of classified-as-healthy patches is higher than a threshold. In order to mitigate the issue of well-chosen training sets, we perform 10 different trials of each experiment with an arbitrary choice of training images. All results reported next are the average of the classification accuracies over 10 trials.
 \par
\begin{figure}[t]
\centering
\subfloat[\label{subfig:udhbases}]{%
  \includegraphics[width=0.24\textwidth]{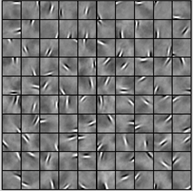}}
~
\subfloat[\label{subfig:dcisbases}]{%
  \includegraphics[width=0.24\textwidth]{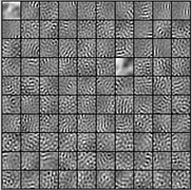}}
 \caption{{Examples of learned bases from \protect\subref{subfig:udhbases} UDH and \protect\subref{subfig:dcisbases} DCIS dictionaries.}}
  \label{fig: ibldictreconstruct}
\end{figure}

Learned bases corresponding to the two classes are visualized in Fig. \ref{fig: ibldictreconstruct}. The average classification accuracy for each method is shown in Table \ref{tab: ibladlresult}. {It is evident from the table that} DFDL outperforms others, offering a classification accuracy of nearly 100 percent in recognizing DCIS and just under 97 percent in UDH. It means that by using DFDL method, the probability of miss is extremely low while that of false alarm is kept at a low level. In order to illustrate the efficiency of DFDL, we keep number of training patches and learned bases but reduce the number of training images by half. Noticeably, DFDL still shows outstanding results compared to the other methods.\par

\textbf{ADL-Lung data set:} This database contains bovine histopathology images of lung acquired by pathologists at the Animal Diagnostics Lab, Pennsylvania State University. These images are scanned using a whole slide digital scanner at 40x optical magnification and are of size $4000\times 3000$ pixels. For the purpose of computational speed-up, all images are downsampled 
to $400\times 300$ pixels in an aliasing-free manner. This database consists of images from two classes: healthy and inflammatory. Each class has 150 images from which 40 images are chosen for training, the remaining ones are used for testing. The averaging experiment results over 10 trials for different methods are presented in Table \ref{tab: ibladlresult}. Apparently, SHIRC and LC-KSVD are moderately suitable to detect inflammatory while WND-CHARM only provides a fairly high-performance in recognizing healthy organs. In contrast, DFDL offers the best accuracy in both detecting Healthy and Inflamed organs with more than 92 percent in the former and over 94 percent in the latter.
\par
\begin{center}
\begin{table}

\caption{ {Classification accuracies: IBL and ADL Lung}}
{\small \vspace{-0.05in}
\label{tab: ibladlresult}
\begin{tabular}{|c||c|c||c|c|}
\hline
\multirow{2}{*}{    Method}          &     \multicolumn{2}{c||}{IBL (\%)}           &  \multicolumn{2}{c|}{ADL Lung (\%)}          \\ \cline{2-5}
          &     UDH  &    DCIS  &   Healthy  &  Inflamed  \\
\hline
\hline
 WND-CHARM \cite{Shamir2008}     &  86.36  &  90.91  &    \textbf{88.75}  &    62.38  \\
 SRC  \cite{Wright2009SRC}    &  68.00  &  56.00  &    72.50  &    75.83  \\
 SHIRC \cite{Srinivas2014SHIRC}   &  93.33  &  90.00  &    75.00  &    85.00  \\
 LC-KSVD \cite{Zhuolin2013LCKSVD} &  58.73  &  54.13  &    77.03  &    83.93  \\
 FDDL \cite{Meng2011FDDL}    &  84.62  &  91.84  &         -  &         -  \\
 DFDL$^{(*)}$     &  \textbf{96.67}  &  \textbf{98.63} &    86.10  &    \textbf{91.28}  \\
 DFDL     &  \textbf{96.94}  & \textbf{ 99.74}  &    \textbf{92.34}  &  \textbf{ 94.24}  \\
 \hline
\end{tabular}}
\begin{tablenotes}
      \small
      \item $^{(*)}$ reduce number of training samples by half.
    \end{tablenotes}
\end{table}
\end{center}
\begin{table}[t]
\centering
{
\caption{ Confusion matrix: MVP and Not MVP}
\label{tab:mvpresult}
{
\begin{tabular}{|c||c|c||c|}
\hline
Class     &  {\small Not MVP(\%)}  &  {\small MVP(\%)}     &  Method           \\
\hline
\hline
 \multirow{2}{*}{\small Not MVP}  &  66.98           &  33.02           &  {\small WND-CHARM}   \cite{Shamir2008}            \\
          &  \textbf{71.70}  &  28.30            &  DFDL  \\
\hline
 \multirow{2}{*}{\small MVP}      &  12.50           &  87.50           &  {\small WND-CHARM}   \cite{Shamir2008}            \\
          &  09.37           &  \textbf{90.63}  &  DFDL  \\
\hline
\end{tabular} }
}
\end{table}
\par
\textbf{TCGA data set:}
In this section, we present {experimental results} on the brain cancer histopathological images obtained from TCGA database \cite{TCGA} provided by the National Institute of Health. One important indicator of a high grade glioma is presence of MicroVascular Proliferation (MVP). Essentially MVP is presence of proliferation of hypertrophic endothelial cells in the tissue. An example of a tissue containing MVP regions is illustrated in Fig. \ref{subfig:MVP}. In this paper, we applied our method to find MVP regions with a slight modification to the decision procedure. This modification is crucial to obtain desirable performance using our algorithm since MVP detection is an inherently more difficult problem because of the complexity of cell structure and morphological features of the cells in MVP regions.

\par Unlike classifying images in IBL and ADL Lung data set which are distinguishable by researching small regions, MVP detection requires more {effort} because an MVP region might be surrounded by tumor cells which are actually low grade. 
We define a patch as \emph{MVP} if it lies entirely within an MVP region and as \emph{Not MVP} otherwise. A new image is divided into non-overlapping patches and it is classified as MVP only if it contains a sufficiently large number of neighboring classified-as-MVP patches.
\par
We use a total of 178 images (resolution {$1800\times 1800$}) {from the TCGA}, 52 for MVP and 126 for not MVP and 20 images are randomly selected from each class for training. We manually extracted MVP and Not MVP regions from training images and randomly extract training patches from these regions for learning.
 Experimental results for DFDL and WND-CHARM are presented in Table \ref{tab:mvpresult}. The Table also shows promising performance of DFDL in detecting MVP  with a accuracy over three percent more than those for state-of-the-art WND-CHARM features.
\newpage
\begingroup
\fontsize{10pt}{10pt}\selectfont
\bibliographystyle{IEEEbib}
  \bibliography{ISBI_2015_revise_2}
\endgroup
\end{document}